\ifdefined\XeTeXversion\else
  \ifdefined\pdfoutput\pdfoutput=1\fi
\fi
\documentclass[10pt, logo, onecolumn, copyright]{nv}

\usepackage[square,sort,comma,numbers]{natbib}
\usepackage{xspace}
\usepackage{bm}
\usepackage{multirow}
\usepackage{placeins}
\usepackage{float}
\usepackage[ruled,vlined,linesnumbered]{algorithm2e}
\usepackage{tikz}
\usetikzlibrary{arrows.meta,positioning,calc,fit,backgrounds}

\usepackage[utf8]{inputenc}
\usepackage[T1]{fontenc}
\usepackage{amsmath,amssymb}
\usepackage[ruled,vlined,linesnumbered]{algorithm2e}
\usepackage{booktabs}
\usepackage{float}
\usepackage{graphicx}
\usepackage{microtype}
\usepackage{xcolor}

\usepackage{graphicx}
\usepackage{amsmath}
\usepackage{hyperref}
\usepackage{url}
\usepackage{enumitem}
\usepackage{multirow}
\usepackage{caption}
\usepackage{listings}
\lstdefinestyle{py}{
  language=Python,
  basicstyle=\ttfamily\small,
  keywordstyle=\color{blue},
  commentstyle=\color{gray}\itshape,
  stringstyle=\color{teal},
  showstringspaces=false,
  breaklines=true,
  numbers=left,
  numberstyle=\tiny\color{gray},
}
\usepackage[ruled,vlined,linesnumbered]{algorithm2e}
\usepackage{amsmath,amssymb}
\usepackage{bm}

\usepackage{subcaption}   %
\usepackage{xspace}       %
\usepackage[capitalise,noabbrev]{cleveref}   %

\setlist[enumerate]{itemsep=0pt, topsep=1pt} %

\hypersetup{
  colorlinks=true,
  urlcolor=nvidiagreen,
  citecolor=nvidiagreen,
  linkcolor=nvidiagreen,
}

\newcommand{\spin}{\textsc{Spin}\xspace}
\newcommand{\dsa}{DSA\xspace}
\newcommand{\dsv}{DeepSeek-V3.2\xspace}
\newcommand{\dsvfour}{DeepSeek-V4\xspace}
\newcommand{\topk}{top-$K$\xspace}
\newcommand{\bench}[1]{\textsc{#1}}

\title{\spin\unskip: Shadow Predictive Indexer for Sparse Attention}
\correspondingauthor{}
\author{%
    Yao~Fu, Jiahan~Chang, Yueying~Li, Ritchie~Zhao, Bryce~Long,\authorcr
    Mohammad~Mahdi~Kamani, Samkit~Jain, Rahul~Raman, Tara~Safavi, Shreya~Gupta,\authorcr
    Parsa~Ashrafi~Fashi, Minseok~Lee, Julien~Demouth, Bita~Darvish~Rouhani%
}

\begin{abstract}

Indexer-based sparse attention reduces the cost of core attention by passing only a fixed, small number of important tokens to it. However, the indexer must still score the entire KV cache at every decoding step. This scoring overhead becomes a major bottleneck as the context length grows.
We propose \spin{} (\textbf{S}hadow \textbf{P}redictive \textbf{In}dexer) to reduce this indexer overhead. \spin{} uses lightweight, history-based prediction to identify important KV blocks, avoiding the need to score the full KV cache at every decoding step. \spin{} keeps the score history fresh with random exploration and allocates the block budget across layers by their measured miss rates. \spin{} treats KV blocks and speculative decoding as first-class design and implementation considerations. Across extensive evaluations on long-context and agentic benchmarks, \spin{} achieves 30--40\% sparsity while preserving task quality. With larger exploration shares and per-layer block budgets, \spin keeps task quality on the especially difficult \bench{MRCRv2} benchmark close to baseline at 50--57\% effective block sparsity. In end-to-end vLLM serving, \spin improves output throughput by up to 13.9\% and reduces median inter-token latency by up to 13.6\%.
\end{abstract}

\begin{document}

\maketitle

\section{Introduction}
\label{sec:introduction}

DeepSeek Sparse Attention~\cite{deepseekv32} (\dsa) is a sparse attention architecture that relies on an indexer to score each token in the context. DSA and its variants have seen significant adoption among frontier open-source models, including \dsvfour~\cite{deepseekv4}, GLM-5.2~\cite{glm52}, MiniMax-M3~\cite{minimaxmsa2026}, and LongCat-2.0~\cite{longcat2026}. Core attention processes a fixed \topk budget, whereas the indexer scales linearly with context length, making it the dominant bottleneck in long-context inference. \dsvfour's compressed sparse attention (CSA) layer~\cite{deepseekv4} combines the original DSA with a new token-level compressor, effectively reducing DSA's context length by $4\times$. This, however, does not change the indexer's linear scaling with context length; thus, at very long context lengths (e.g., 512K--1M), the indexer remains a major contributor to \dsvfour's decode inter-token latency (ITL).

We observe that DSA's indexer patterns are predictable, similar to temporal patterns previously observed in attention scores~\cite{attnpredictor2025,tappa2026}. Indexer scores can be positively correlated with (1) previous DSA layers' indexer scores in the same decode iteration (\textit{prev-layer}) and (2) the same layer's scores from previous iterations (\textit{prev-iter}). Our benchmarking results show that prev-iter signals correlate more strongly with current scores than prev-layer signals. This motivates \textbf{S}hadow \textbf{P}redictive \textbf{In}dexer (\textbf{SPIN}), which uses simple per-request statistics of each DSA layer's scores from previous decoding iterations to predict token importance and skip less important context tokens, thereby sparsifying the indexer input. Because skipped blocks receive no score updates, \spin reserves a small share of a fixed block budget for random exploration, which refreshes them at low overhead. Indexer scores are also not equally predictable across layers, so \spin sizes a \emph{per-layer block budget} by each layer's miss rate, measured at periodic \emph{dense probe steps}. To enable efficient implementation in LLM serving frameworks, \spin applies sparsity at the KV-block level. \spin is training-free and introduces no new auxiliary layers or networks.

We implement \spin for the \dsvfour model class in vLLM~\cite{vllm2023} for accuracy evaluation. Across the long-context benchmarks \bench{LongBench-v2}~\cite{bai2025longbenchv2deeperunderstanding}, \bench{AA-LCR}~\cite{artificialanalysis2025lcr}, \bench{MRCRv2}~\cite{openai2025mrcr}, and \bench{RULER}~\cite{ruler2024}, and the agentic benchmark \bench{tau2-airline}~\cite{yao2024tau,barres2025tau2}, \spin preserves comparable task quality at 30--40\% block sparsity. On \bench{MRCRv2}, scaling the exploration share with target block sparsity keeps task quality close to baseline at 50\% block sparsity, and per-layer block budgets reach 57\% effective block sparsity with the default 5\% share. We further integrate \spin into vLLM, where the 57\% setting improves output throughput by 13.9\% and reduces median inter-token latency by 13.6\%.

Our concrete contributions are as follows:
\begin{enumerate}[nosep]
    \item We evaluate various prev-iter and prev-layer prediction signals for the indexer in \dsvfour and show that prev-iter indexer scores provide the most accurate predictions.
    \item We propose \spin to sparsify the indexer based on prev-iter scores and an exploration mechanism and ablate various methods to pool the per-token scores into a block-level prediction.
    \item We compare exploration methods on \bench{MRCRv2} and show that exploration keeps task quality close to baseline within the same block budget when the exploration share grows with target block sparsity.
    \item We size \spin's block budget per layer with a min-max partition of per-layer miss curves measured from periodic dense probe steps, which keeps \bench{MRCRv2} task quality close to baseline at 57\% effective block sparsity.
    \item We evaluate \spin on \dsvfour across long-context and agentic benchmarks and integrate it into vLLM, improving output throughput by 13.9\% and reducing median inter-token latency by 13.6\% at 57\% effective block sparsity.
    \item We evaluate \spin during speculative target-model verification using \dsvfour's native multi-token prediction (MTP) on \bench{SPEED-Bench}~\cite{speedbench2026}.
\end{enumerate}

\section{DSA Indexer Patterns and Prediction}
\label{sec:patterns}

In each decode iteration, DSA's indexer computes a scalar score for each context token.
For query token $t$, the indexer projects its representation into a multi-head query $q_t$. The key $k_i$ for each context token $i$ is generated earlier and cached during inference.
Rotary positional encodings (RoPE)~\cite{rope2021} are applied to half of each head's dimensions.
The indexer score for context token $i$ is $z_t[i] = \sum_{h=1}^{n_h} w_h \, \text{ReLU}(q_{t,h} \cdot k_{i,h})$,
where $n_h$ is the number of heads, and $q_{t,h}$, $k_{i,h}$, and $w_h$ are the query, key, and weight for head $h$.
The \topk tokens with the highest scores are selected and passed to core attention,
with $K$ in practice ranging from $512$ to $2048$.

\begin{figure*}[!htbp]
    \centering
    \begin{minipage}{0.49\textwidth}
        \centering
        \dsv Indexer \topk \\
        \includegraphics[width=\textwidth,
            trim=0pt -10pt 0pt 28pt, clip]{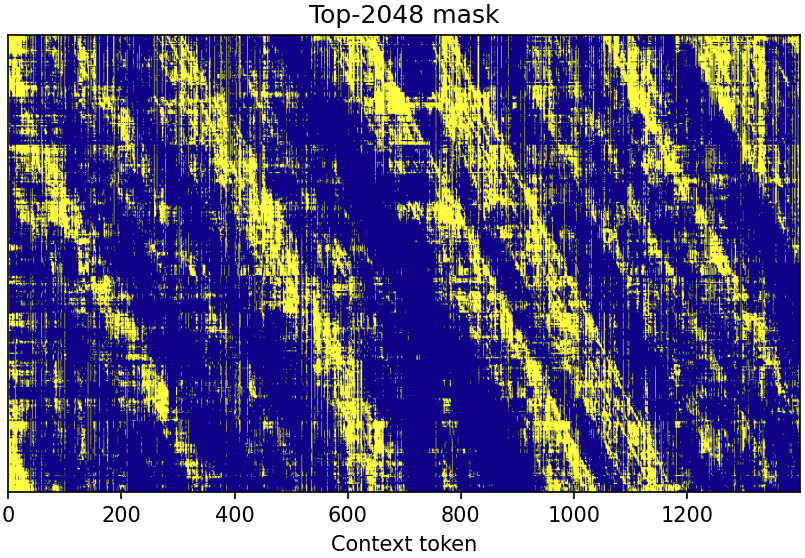}
    \end{minipage}
    \begin{minipage}{0.49\textwidth}
        \centering
        \dsvfour Indexer \topk \\
        \includegraphics[width=\textwidth,
            trim=0pt -10pt 0pt 28pt, clip]{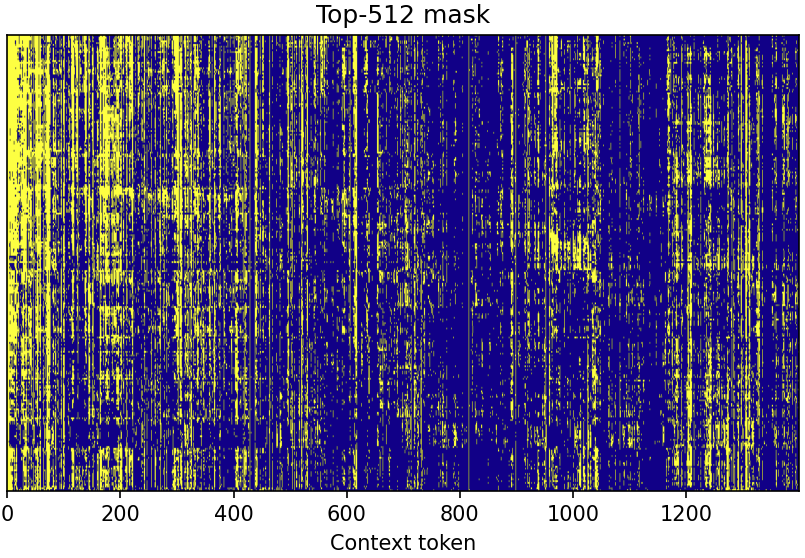}
    \end{minipage}
    \caption{\textbf{Indexer \topk patterns} from a single DSA layer on a prompt from \bench{summscreenfd}.
        The y-axis represents decode iterations. The prompt is 9.7K tokens long, and 1,400 context KV slots are visualized along the x-axis.
        \dsv exhibits prominent diagonal bands and vertical lines, whereas \dsvfour exhibits mostly vertical patterns.
    }
    \label{fig:indexer-patterns}
\end{figure*}

The DSA indexer exhibits patterns that can be predicted in different ways.
Figure~\ref{fig:indexer-patterns} visualizes the \topk selection mask from a single layer on a prompt from \bench{summscreenfd}~\cite{summscreen2022}.
The vertical axis represents decode iterations. Vertical-line and diagonal-band patterns are clearly visible.
Vertical lines indicate that the indexer tends to select the same tokens across multiple iterations.
Diagonal bands resemble well-documented "slash" patterns in some attention heads~\cite{slashrope2026,tappa2026},
which are known to be caused by RoPE~\cite{barbero2025round,gvr2026}.
Among the DSA models we examine, \dsv is the only one with prominent diagonal bands; the others exhibit mostly vertical patterns.
Both patterns allow prediction of indexer scores using scores from the previous iteration (\textit{prev-iter} prediction).

Indexer scores can also be predicted using scores from earlier layers in the same decode iteration (\textit{prev-layer} prediction).
This is the core idea behind GLM-5's IndexCache~\cite{indexcache2026} and indexer sharing across adjacent layers in GLM-5.2~\cite{glm52}.
We compare four predictors derived from two signal sources---previous layers and previous decode iterations---against a random baseline. Each method constructs a proxy for the target indexer scores:
\begin{enumerate}
    \item \textbf{Random}: Sample indexer scores independently and uniformly at random. The expected score correlation is zero.
    \item \textbf{Prev-layer}: Reuse the indexer scores from the preceding DSA layer in the same decode iteration.
    \item \textbf{Prev-layer (best-of-5 oracle)}: Consider the indexer scores from the five preceding DSA layers and report the best of the five.
    \item \textbf{Prev-iter}: Reuse the indexer scores from the preceding decode iteration at the same DSA layer.
    \item \textbf{Prev-iter (offset 1)}: Shift the prev-iter indexer scores by one position to align with the diagonal pattern.
\end{enumerate}

To test whether the prediction signals generalize across models, tasks, and context lengths, we evaluate them on both \dsv and \dsvfour Flash using two diagnostic prompts: a 9.7K-token \bench{summscreenfd} summarization prompt and a 94K-token \bench{AA-LCR}~\cite{artificialanalysis2025lcr} reasoning prompt. At each decode iteration, we compare the predicted scores and \topk indices with the original indexer outputs, reporting Pearson correlation for score prediction and recall for \topk selection. For \bench{summscreenfd}, we disable thinking and evaluate all DSA layers. For \bench{AA-LCR}, we enable thinking but evaluate only layers $\{0,10,20,30,40,50,60\}$ and omit the prev-layer best-of-5 oracle to limit evaluation cost.

\begin{table*}[!htb]
    \centering
    \caption{\textbf{Comparison of indexer prediction methods} --- profiling results for one prompt each from
        \bench{summscreenfd} (length: 9.7k) and \bench{AA-LCR} (length: 94k).
        Values are averaged over layers and decode iterations.
        The best method is shown in bold, and the second-best method is underlined.
    }
    \label{tab:predict-comparison}
    \begin{tabular*}{\textwidth}{l@{\extracolsep{\fill}}rrrrrrrr}
\toprule
& \multicolumn{4}{c}{\textbf{DeepSeek-V3.2}}
& \multicolumn{4}{c}{\textbf{DeepSeek-V4 Flash}} \\
\cmidrule(lr){2-5}
\cmidrule(lr){6-9}
& \multicolumn{2}{c}{\texttt{summscreenfd}}
& \multicolumn{2}{c}{\texttt{AA-LCR}}
& \multicolumn{2}{c}{\texttt{summscreenfd}}
& \multicolumn{2}{c}{\texttt{AA-LCR}} \\
\cmidrule(lr){2-3}
\cmidrule(lr){4-5}
\cmidrule(lr){6-7}
\cmidrule(lr){8-9}
Method
& \shortstack{Pearson\\correlation} & \shortstack{\topk\\recall}
& \shortstack{Pearson\\correlation} & \shortstack{\topk\\recall}
& \shortstack{Pearson\\correlation} & \shortstack{\topk\\recall}
& \shortstack{Pearson\\correlation} & \shortstack{\topk\\recall} \\
\midrule
Random baseline      & 0.000          & 0.203          & 0.000          & 0.022          & 0.000          & 0.203          & 0.000          & 0.022          \\
Prev-layer           & 0.289          & 0.359          & 0.431          & 0.164          & 0.547          & 0.482          & 0.519          & \underline{0.306} \\
Prev-layer (Bo5)     & 0.500          & 0.461          & --             & --             & \underline{0.626} & \underline{0.525} & --            & --             \\
Prev-iter            & \textbf{0.913} & \textbf{0.824} & \textbf{0.902} & \textbf{0.615} & \textbf{0.876} & \textbf{0.751} & \textbf{0.849} & \textbf{0.577} \\
Prev-iter (offset 1) & \underline{0.807} & \underline{0.660} & \underline{0.665} & \underline{0.314} & 0.526 & 0.460 & \underline{0.571} & 0.302 \\
\bottomrule
\end{tabular*}

\end{table*}

Table~\ref{tab:predict-comparison} shows the score correlation and \topk recall for each prediction method, averaged over layers and decode iterations. Prev-iter prediction is the most accurate method and also scales well to longer contexts. From \bench{summscreenfd} to \bench{AA-LCR}, the context length increases by a factor of $9.7$, and the random recall baseline drops proportionally. However, the prev-iter recall rate is largely maintained. Prev-layer prediction works better on \dsvfour than on \dsv but lags prev-iter on both models.

Motivated by these results, we design \spin to predict block importance from per-layer statistics accumulated across previous decode iterations.

\section{SPIN}
\label{sec:method}

\subsection{Overview}

\spin extends naive prev-iter score prediction in four ways:
\begin{enumerate}
    \item Maintain exponential moving averages (EMAs) of previous indexer scores along the vertical and diagonal patterns. This allows \spin to capture longer-term signals in both patterns.
    \item Pool the predicted token-level scores into blocks. This allows \spin to skip entire KV blocks and integrate efficiently with paged-attention backends.
    \item Reserve a small share of the block budget for random exploration. This allows \spin to refresh observations of skipped blocks without increasing the number of blocks scored.
    \item Allocate a per-layer block budget by each layer's miss rate, measured at periodic dense probe steps. Layers with higher miss rates receive more blocks.
\end{enumerate}

\subsection{\spin Algorithm}

\spin operates over the KV slots visible to the indexer; for \dsvfour, each slot corresponds to a compressed token.

To predict the indexer score of slot $i$ at query position $q_t$, \spin combines two EMA statistics: $V[i]$, the vertical score associated with absolute slot $i$, and $D[q_t-i]$, the diagonal score associated with relative offset $q_t-i$. The predicted score is
\begin{equation}
    \hat{z}_t[i] = \max\left(V[i], D[q_t-i]\right).
\end{equation}

\spin partitions the slots into logical blocks of size $b$, matching the block size used by the attention backend, and defines $b(i)=\lfloor i/b\rfloor$ as the block containing slot $i$. It then max-pools the predicted scores within each block to produce a predicted block score:
\begin{equation}
    S_t[j] = \max_{i:\,b(i)=j} \hat{z}_t[i].
\end{equation}
We study alternative pooling methods in Section~\ref{sec:block-reduction}.

Let $K$ denote the output size of the post-indexer top-$K$ selection. The target block sparsity $\rho\in[0,1]$ denotes the desired fraction of context blocks to skip. The resulting number of selected blocks, $m_t$, is lower-bounded by $\lceil K/b\rceil$ to provide capacity for at least $K$ slots and capped by the number of available blocks. The $m_t$ highest-scoring blocks are passed to the indexer, which computes scores for all slots in those blocks. The original post-indexer top-$K$ operator then selects $K$ slots from these scores for core attention. The pre-indexer block selection operates over one predicted score per block rather than one per slot, reducing the size of the selection problem.

After the indexer evaluates the selected blocks, their scores update the corresponding $V$ and $D$ EMA states. Because skipped blocks produce no scores, \spin updates its EMA state from partial observations at each decode iteration. Slots in skipped blocks keep their previous statistics until they are observed again, and exploration (Section~\ref{sec:exploration}) keeps these observations fresh.

By design, \spin's block prediction and EMA update can be scheduled asynchronously and overlapped with the forward pass of other layers.

\begin{algorithm}[t]
\caption{Core \spin block-selection and EMA-update logic for one decode iteration}
\label{alg:a1-max}
\KwIn{Indexer query position $q_t$; context length $N_t$ in KV slots; state $V,D$;
      block size $b$; post-indexer selection size $K$; target block sparsity $\rho$;
      EMA decay $\alpha$}
\KwOut{Post-indexer top-$K$ slot set $T_t$ and updated state $V,D$}
$B_t \gets \lceil N_t/b\rceil$ \tcp*{total number of context blocks}
\For{$j\gets 0$ \KwTo $B_t-1$}{
    $S_t[j] \gets \max_{jb\le i<\min\{(j+1)b,N_t\}}
        \max\{V[i],D[q_t-i]\}$ \tcp*{predicted scores}
}
$m_t \gets \min\!\left\{B_t,
    \max\!\left(\lceil K/b\rceil,\lceil(1-\rho)B_t\rceil\right)\right\}$ \tcp*{number of blocks to select}
$M_t^B \gets \operatorname{TopK}(\{S_t[0],\ldots,S_t[B_t-1]\},m_t)$    \tcp*{selected blocks}
$z_t \gets \operatorname{Indexer}(M_t^B)$ \tcp*{indexer scores in logical KV-slot coordinates}
$T_t \gets \operatorname{TopK}(z_t,K)$\;
\ForEach{$i\in \operatorname{dom}(z_t)$}{
    $V[i] \gets \alpha V[i]+(1-\alpha)z_t[i]$  \tcp*{EMA updates}
    $D[q_t-i] \gets \alpha D[q_t-i]+(1-\alpha)z_t[i]$\;
}
\end{algorithm}

Algorithm~\ref{alg:a1-max} summarizes \spin's core block-selection and EMA-update logic. For clarity, it omits several practical details. \spin resets $V$ and $D$ to zero after prefill. For multi-turn inference, it repeats this reset at the start of each turn. In both cases, the first decode iteration runs without block pruning to initialize the EMA state. We also always retain the sink block and the newest active block. Exploration is described separately below.

\subsubsection{Exploration}
\label{sec:exploration}
Blocks outside the selection receive no EMA updates, yet they can become important at later decode steps. \spin refreshes them within the existing block budget by reserving part of it for exploration.

At each iteration, \spin reserves a fraction $\gamma$ of the block budget $m_t$ for exploration. We call $\gamma$ the \emph{exploration share} and use 5\% by default. The remaining budget retains the forced blocks and the highest-scoring blocks following Algorithm~\ref{alg:a1-max}. Exploration fills the reserved positions with distinct blocks chosen from those not already selected. Forced blocks count toward $m_t$, and the exploration budget is capped by the remaining capacity. The total number of selected blocks remains unchanged, preserving the block budget.

The explored blocks receive fresh indexer scores and EMA updates, allowing previously missed slots to reflect their current importance. \spin uses random exploration, which needs no per-block state. Section~\ref{sec:staleness} compares exploration methods and exploration shares.

\subsubsection{Per-Layer Block Budgets}
\label{sec:per-layer-budget}
\begin{figure}[!ht]
    \centering
    \includegraphics[width=\textwidth]{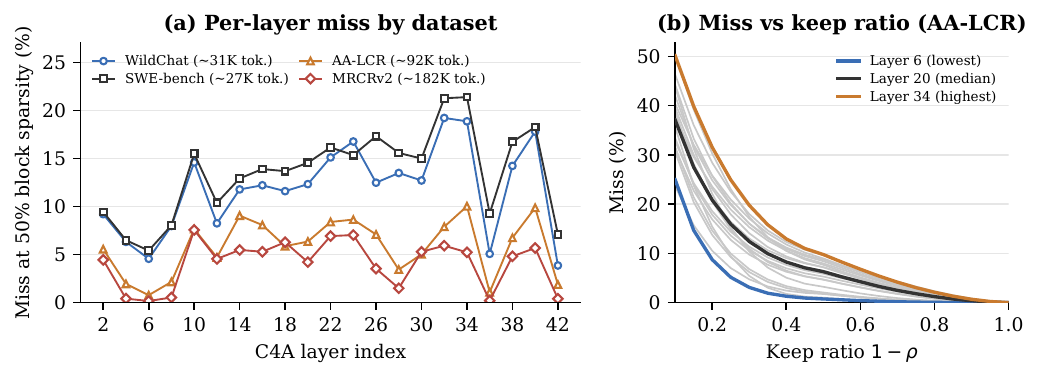}
    \caption{Per-layer miss rate of block prediction for \dsvfour Flash (FP8, GB300 TP4), from dense probe steps while decoding 16 prompts from each of \bench{WildChat}~\cite{wildchat2024}, \bench{SWE-bench}~\cite{swebench2024}, \bench{AA-LCR}, and \bench{MRCRv2}. (a) Miss rate per CSA layer with the same 50\% block sparsity in every layer; the legend gives each prompt set's median length. Pairwise Spearman correlations of the layer ordering across prompt sets range from 0.59 to 0.94. (b) Miss rate versus per-layer keep ratio on \bench{AA-LCR}; the layers with the lowest, median, and highest miss at keep ratio 0.5 are highlighted.}
    \label{fig:per-layer-miss}
\end{figure}

Indexer scores are not equally predictable across layers. With the same 50\% block sparsity in every layer, the \emph{miss rate}, the fraction of the dense indexer's \topk slots that fall into skipped blocks, ranges from 0.8\% to 10.0\% across the 21 CSA layers, the only layers with an indexer, on \bench{AA-LCR}. The layer ordering is similar across the four prompt sets we probe (Figure~\ref{fig:per-layer-miss}). A uniform budget would leave the least predictable layers short of blocks, so \spin sizes the budget per layer. A \emph{per-layer block budget} lets each CSA layer $l$ keep its own fraction $k_l$ of context blocks, its \emph{keep ratio}, subject to a fixed total budget
\begin{equation*}
\textstyle\sum_l k_l \le L(1-\rho), \qquad k_l \in \mathcal{G}=\{k_{\min}, k_{\min}+\delta, \ldots, 1\},
\end{equation*}
where $L$ is the number of CSA layers. Each layer's indexer cost is close to linear in $k_l$, so total indexer work depends only on the mean keep ratio.

Every $P$ decode steps, where $P$ is the \emph{probe period}, a \emph{dense probe step} scores all blocks. Before the EMA update, \spin records, for every keep ratio $k \in \mathcal{G}$, the miss rate of its top-ranked $\lceil k B_t\rceil$ predicted blocks. This gives each layer's \emph{miss curve} $\mu_l(k)$ at the cost of one dense indexer step.

\spin chooses the keep ratios that minimize the worst layer's miss, $\min \max_l \mu_l(k_l)$, under the budget constraint. This \emph{min-max partition} has an exact threshold-scan solution. For a miss threshold $\tau$, each layer takes the smallest keep ratio that meets it, and \spin takes the smallest threshold that fits the budget,
\begin{equation*}
k_l(\tau)=\min\{k\in\mathcal{G}:\mu_l(k)\le\tau\}, \qquad
\tau^\star=\min\Big\{\tau\in\{\mu_l(k)\}:\textstyle\sum_l k_l(\tau)\le L(1-\rho)\Big\}.
\end{equation*}
Because $\sum_l k_l(\tau)$ never increases with $\tau$, one scan over the measured miss values finds $\tau^\star$. \spin spends any leftover budget one grid step $\delta$ at a time on the layer with the highest current miss, and the floor $k_{\min}$ bounds how far any layer can be starved. Each layer $l$ then runs Algorithm~\ref{alg:a1-max} with its own target block sparsity $1-k_l$.

A background thread re-solves the partition every 60\,s when any miss curve drifts by more than 0.02, and publishes the keep ratios to a device-side table that works under CUDA-graph replay without recompilation. Probe steps skip no blocks, so we report per-layer block budgets at their \emph{effective block sparsity} $\rho(1-1/P)$. We use $\delta=0.05$, $k_{\min}=0.1$, and the default 5\% exploration share. A \spin configuration is set by $\rho$, $\gamma$, and $P$. Without probe steps, every layer keeps $1-\rho$.

\subsubsection{Speculative Decoding}
For a speculative draft of length $d$, the target model evaluates $d+1$ verifier queries in a single forward pass, indexed by rows $r\in\{0,\ldots,d\}$. Row $r$ is valid if the first $r$ draft tokens are accepted; row 0 is therefore always valid. These verifier queries share the same context KV cache but have different query positions and hidden states.

A potential concern when applying \spin to speculative decoding is that independently predicted block masks for each query may have a large union, reducing the realized block sparsity. However, we observe that the block masks selected for adjacent verifier queries overlap substantially (Section~\ref{sec:mtp-results}). Based on this observation, \spin predicts a block mask for each verifier query and uses their union as the shared indexer-input mask.

EMA updates are deferred until the verification outcomes are known. If $a$ draft tokens are accepted, the observations from rows $0,\ldots,a$ are replayed in order to update the EMA state. The remaining rows are off the committed decode path and are discarded.

We evaluate this design using \dsvfour's native MTP implementation and study its impact on speculative-decoding acceptance in Section~\ref{sec:mtp-results}.

\section{Experimental Evaluation}
\label{sec:eval}

In this section, we evaluate \spin's impact on inference quality and performance. We also conduct microbenchmarks of its key design choices. Specifically, we answer the following questions:

\begin{enumerate}
    \item How does \spin affect task quality on long-context and agentic benchmarks?
    \item How much does \spin improve end-to-end output throughput and inter-token latency in vLLM?
    \item How does \spin interact with speculative decoding, and does applying it during target-model verification affect average acceptance length?
    \item How do block pooling, exploration, and per-layer block budgets contribute to \spin's selection accuracy and task quality?
\end{enumerate}

\subsection{Experimental Setup}

We evaluate \spin on \dsvfour Flash and Pro, using Flash as the primary model. The evaluation covers long-context benchmarks (\bench{LongBench-v2}, \bench{AA-LCR}, \bench{MRCRv2}, and \bench{RULER}), the agentic benchmark \bench{tau2-airline}, and speculative decoding on \bench{SPEED-Bench}. Unless otherwise specified, sparse experiments use logical blocks of 64 compressed indexer slots, max block pooling, and an EMA decay of $0.5$. The LongBench-v2, AA-LCR, RULER, and tau2-airline results run \spin without exploration, which already keeps their task quality comparable to all-keep (Section~\ref{sec:long-context}). \bench{MRCRv2} is where exploration matters (Section~\ref{sec:mrcr}). The block-reduction and MTP results also run without exploration. The MRCRv2 study compares exploration methods and exploration shares. Unless a table lists the exploration share, results with exploration, including the end-to-end implementation, use random exploration with a 5\% share. Results use the same block budget in every layer unless they list a probe period $P$. Per-layer results report effective block sparsity, the target block sparsity with every probe step counted as dense (Section~\ref{sec:per-layer-budget}). We use \spin~($x\%$) to denote a target block sparsity of $x\%$, meaning that $x\%$ is the intended fraction of context blocks to skip. Realized block sparsity is the measured fraction skipped after budget rounding, forced-block retention, and, for speculative decoding, mask union. We compare each \spin configuration with a matched \emph{all-keep} baseline, in which no blocks are skipped, using the same model and evaluation protocol.

To measure mean and p99 miss, we compute dense indexer scores within the \dsvfour indexer path and apply \spin's block masks before \topk selection. As in sparse execution, \spin updates its EMA states only from the selected blocks. We report the task-specific score as the primary quality metric and \topk miss rate (mean and p99) as auxiliary diagnostic metrics. The miss rate is the fraction of ground truth \topk indices removed due to \spin's block-sparse mask. In the MRCRv2 exploration study, the indexer scores only the selected blocks, and we report task scores. Benchmark-specific setup details and hyperparameters are described in their respective subsections. End-to-end serving performance is measured using the integrated vLLM implementation described in Section~\ref{sec:end-to-end-performance}.

\subsection{Long-context Benchmarks}
\label{sec:long-context}
\paragraph{Cross-model quality.}
We evaluate whether \spin preserves task quality on \dsvfour Flash and Pro using \bench{LongBench-v2}. We follow the official evaluation protocol but increase the maximum output length from 128 to 1,024 tokens to reduce output truncation. Table~\ref{tab:lbv2} reports the zero-shot exact-match scores for all-keep and \spin configurations. Across target sparsities from 20\% to 50\%, all \spin scores differ from the corresponding all-keep score by at most 2.6\% in relative terms. These results show that \spin maintains comparable task quality on both Flash and Pro with up to 50\% target block sparsity.

\begin{table}[h]
    \centering
    \caption{LongBench-v2 exact-match score (\%) over all 503 examples. \spin runs without exploration.}
    \label{tab:lbv2}
    \begin{tabular}{lrr}
        \toprule
        Configuration & \dsvfour Flash & \dsvfour Pro \\
        \midrule
        All-keep     & 53.48 & 58.05 \\
        \spin~(20\%) & 53.08 & 57.85 \\
        \spin~(30\%) & 52.09 & 57.06 \\
        \spin~(40\%) & 54.27 & 58.25 \\
        \spin~(50\%) & 54.08 & 58.65 \\
        \bottomrule
    \end{tabular}
\end{table}

\paragraph{Sensitivity to selection misses.}
We evaluate how sensitive task quality is to the selection misses introduced by \spin on \bench{AA-LCR} and \bench{RULER} using \dsvfour Flash. The \bench{AA-LCR} evaluation contains 100 questions with mean and maximum prompt lengths of 95.5K and 115.2K tokens, respectively. We repeat the evaluation five times with different random seeds to obtain a more stable score estimate. \bench{RULER} covers 13 tasks at six context lengths from 4K to 128K, with 500 examples per task--length cell, yielding 39,000 examples per configuration. Given this large evaluation set, we run each configuration once.

\begin{table}[h]
    \centering
    \caption{AA-LCR and RULER results for \dsvfour Flash. AA-LCR scores are averaged over execution-level runs; RULER scores are macro-averaged over task--length cells. \spin runs without exploration.}
    \label{tab:long-context-diag}
    \begin{tabular*}{\linewidth}{l@{\extracolsep{\fill}}rrrrrr}
        \toprule
        & \multicolumn{3}{c}{AA-LCR} & \multicolumn{3}{c}{RULER} \\
        \cmidrule(lr){2-4}
        \cmidrule(lr){5-7}
        Configuration & Score & Mean miss & P99 miss & Score & Mean miss & P99 miss \\
        \midrule
        All-keep     & 64.40 & --    & --     & 90.20 & --     & --     \\
        \spin~(20\%) & 62.60 & 0.88\% & 7.81\%  & 90.13 & 5.72\%  & 19.92\% \\
        \spin~(30\%) & 64.60 & 1.92\% & 13.48\% & 90.00 & 10.07\% & 29.49\% \\
        \spin~(40\%) & 64.00 & 3.46\% & 20.31\% & 89.69 & 14.87\% & 38.67\% \\
        \spin~(50\%) & 62.00 & 5.78\% & 28.52\% & 88.96 & 20.68\% & 48.83\% \\
        \bottomrule
    \end{tabular*}
\end{table}

Table~\ref{tab:long-context-diag} reports the task score, mean miss, and p99 miss for all-keep and different sparse configurations. Across target sparsities from 20\% to 50\%, \bench{AA-LCR} scores differ from all-keep by at most 2.4 points, or 3.7\% in relative terms, without a monotonic decrease. On \bench{RULER}, scores decrease slightly and monotonically as miss rates increase. \bench{RULER} exhibits substantially higher miss rates than \bench{AA-LCR}: at 50\% sparsity, its p99 miss reaches 48.83\%, but task-score degradation remains small---only 1.24 points, or 1.4\% relative to all-keep.

\subsection{MRCRv2 Stress Test}
\label{sec:mrcr}
We evaluate \spin on \dsvfour Flash using the 8-needle \bench{MRCRv2} setting at 256K, 512K, and 1M context lengths. Each configuration uses the same 100 official prompts per context bin over 40 seeded repeats, with thinking enabled at maximum reasoning effort and an output limit of 8,192 tokens. We serve the model with vLLM v0.31.0 on four GB300 GPUs (TP4) with FP8 KV and indexer caches, batch size 4, 16K-token chunked prefill, and prefix caching. All-keep uses the same deployment with \spin disabled. The three examples that exceed the model context limit in the 1M bin retain zero scores. Table~\ref{tab:mrcr-basic} reports \spin with its default 5\% exploration share and, as an ablation, without exploration.

\begin{table*}[h]
    \centering
    \caption{MRCRv2 task scores of \spin with the same block budget in every layer, with 5\% random exploration and, as an ablation, without exploration. Each score averages 40 repeats over 100 fixed prompts per context bin. The 95\% confidence intervals of the paired difference from all-keep span $\pm$0.9 to $\pm$1.5 points.}
    \label{tab:mrcr-basic}
    \begin{tabular*}{\textwidth}{l@{\extracolsep{\fill}}rrrrrr}
        \toprule
        & \multicolumn{2}{c}{256K}
        & \multicolumn{2}{c}{512K}
        & \multicolumn{2}{c}{1M} \\
        \cmidrule(lr){2-3}
        \cmidrule(lr){4-5}
        \cmidrule(lr){6-7}
        \shortstack[l]{Target block\\sparsity} & Random & No exploration & Random & No exploration & Random & No exploration \\
        \midrule
        All-keep & \multicolumn{2}{c}{74.67} & \multicolumn{2}{c}{59.57} & \multicolumn{2}{c}{43.19} \\
        \midrule
        20\% & 75.38 & 73.51 & 58.93 & 58.41 & 42.66 & 42.13 \\
        30\% & 73.98 & 70.75 & 59.61 & 53.84 & 42.37 & 40.15 \\
        40\% & 73.29 & 63.75 & 56.99 & 48.55 & 41.39 & 36.90 \\
        50\% & 70.31 & 54.90 & 54.77 & 44.09 & 39.88 & 33.93 \\
        \bottomrule
    \end{tabular*}
\end{table*}

With a 5\% exploration share, \spin keeps task quality close to all-keep at 20--30\% block sparsity across all three context lengths, and a 30\% share keeps it close up to 50\% (Section~\ref{sec:staleness}). Exploration matters most on this stress test. Without it, skipped blocks keep stale scores, and \bench{MRCRv2} task quality falls as block sparsity grows. Section~\ref{sec:staleness} shows how the share grows with block sparsity.

\subsection{Agentic Benchmarks}
\paragraph{\bench{tau2-airline}}

We evaluate \spin with \dsvfour Flash on \bench{tau2-airline}, a multi-turn agentic benchmark containing 50 tasks. We run four seeded trials per task for each configuration, yielding 200 trajectories per configuration. The target agent uses temperature~0 and top-$p$~1. The hosted GPT-5.2 user simulator is fixed across all configurations at temperature~0.

\begin{table}[h]
    \centering
    \caption{\texttt{tau2-airline} results for DeepSeek-V4 Flash. Mean and p99
    miss are selection-fidelity metrics over eligible sparse decode
    rows and are not applicable to the all-keep control. \spin runs without exploration.}
    \label{tab:tau2-airline}
    \begin{tabular}{lrrr}
        \toprule
        Agent configuration & Score & Mean miss & P99 miss \\
        \midrule
        All-keep        & 72.5\% & --    & --    \\
        \spin~(20\%)    & 70.5\% & 5.9\% & 21.1\% \\
        \spin~(30\%)    & 73.5\% & 10.9\% & 30.9\% \\
        \spin~(40\%)    & 70.0\% & 17.3\% & 41.4\% \\
        \spin~(50\%)    & 69.5\% & 25.7\% & 53.1\% \\
        \bottomrule
    \end{tabular}
\end{table}

At 30\% target sparsity, \spin shows no degradation in task score. Even at 50\% sparsity, the score decreases by only 4.1\% relative to all-keep, despite the mean miss reaching 25.7\%.

In a separate evaluation on \bench{Terminal-Bench 2.1} with \dsvfour Flash, \spin at 40\% target sparsity matches the all-keep score (62.5\% for both).

\subsection{End-to-End vLLM Performance}
\label{sec:end-to-end-performance}

We integrate \spin into vLLM v0.31.0 and evaluate its end-to-end serving performance with \dsvfour Flash on eight NVIDIA B300 GPUs using tensor parallelism of eight (TP8) and expert parallelism of eight (EP8). The routed experts use NVFP4 weights and activations. Attention, the shared experts, the LM head, and the KV and indexer caches use FP8. Serving uses full and piecewise CUDA graphs with MTP disabled. These runs use the same block size, max pooling, EMA decay, warmup, and forced-block policies as the task-quality evaluations. Table~\ref{tab:end-to-end-performance} lists the exploration share of each configuration, and Sections~\ref{sec:mrcr} and~\ref{sec:staleness} report \bench{MRCRv2} task quality for each row.

We evaluate a workload with a long cached context, a common setting in agentic tasks. We pre-populate one distinct cached prefix for each request before sending the measured wave. The workload contains 507K cached context tokens and 5K new input tokens. After one warmup batch, we measure 96 concurrent requests, each generating 16K output tokens. We report output throughput in tokens per second (TPS) and median inter-token latency (ITL).

\begin{table*}[h]
    \centering
    \caption{End-to-end serving performance for \dsvfour Flash with 507K cached context tokens and 5K new input tokens. The batch size is 96, and each request generates 16K tokens. TPS gain and ITL reduction are computed relative to the all-keep row. All rows come from one measurement run. Rows without a probe period use the same block budget in every layer.}
    \label{tab:end-to-end-performance}
    \begin{tabular*}{\textwidth}{l@{\extracolsep{\fill}}rrrrrrr}
        \toprule
        \shortstack{Target block\\sparsity} & \shortstack{Exploration\\share} & \shortstack{Probe\\period} & \shortstack{Effective block\\sparsity} & \shortstack{Output\\TPS} & \shortstack{TPS\\gain} & \shortstack{Median ITL\\(ms)} & \shortstack{ITL\\reduction} \\
        \midrule
        0\% (all-keep) & -- & -- & -- & 4788.86 & -- & 17.777 & -- \\
        30\% & 5\% & -- & 30\% & 5000.56 & +4.4\% & 16.935 & 4.7\% \\
        40\% & 20\% & -- & 40\% & 5192.38 & +8.4\% & 16.278 & 8.4\% \\
        50\% & 20\% & -- & 50\% & 5362.07 & +12.0\% & 15.594 & 12.3\% \\
        65\% & 5\% & 8 & 56.9\% & 5455.49 & +13.9\% & 15.355 & 13.6\% \\
        \bottomrule
    \end{tabular*}
\end{table*}

Table~\ref{tab:end-to-end-performance} reports the end-to-end results. \spin improves output throughput by 4.4--13.9\% and reduces median ITL by 4.7--13.6\%. Throughput gain grows steadily with block sparsity, so sparsity that \spin keeps at high quality turns directly into serving speed. The last row uses per-layer block budgets (Section~\ref{sec:per-layer-budget}), refit live during the run, and its throughput gain follows its effective block sparsity. The exploration share does not change throughput. In the same measurement run, raising it from 20\% to 30\% at 50\% target block sparsity changes output TPS by 0.05\%.

\subsection{Speculative Decoding}
\label{sec:mtp-results}

We study how \spin interacts with speculative decoding using \dsvfour's native multi-token prediction (MTP) with a draft length of three. Extending \spin to multiple verifier queries requires decisions about constructing a shared sparse mask and updating the EMA state after verification. We first ablate these choices through offline trace replay and then evaluate the selected policy's impact on speculative-decoding acceptance online.

\paragraph{Offline policy ablation.}
We collect 60 MTP traces by running 15 curated \bench{AA-LCR} prompts with \dsvfour Flash and Pro in both chat and thinking modes. Each trace records the verifier queries and their acceptance outcomes.

For mask construction, we compare row-0 sharing, which shares the row-0 mask across all verifier rows, with row-local union, which unions masks predicted separately for each row. For EMA updating, we compare row-0 update with prefix-valid update, which sequentially updates from all prefix-valid rows after the acceptance outcomes are known. We replay \spin on the collected traces at seven target block sparsities from 10\% to 50\% and report the aggregated p99 miss rate.

\begin{figure*}[t]
    \centering
    \includegraphics[width=\textwidth]{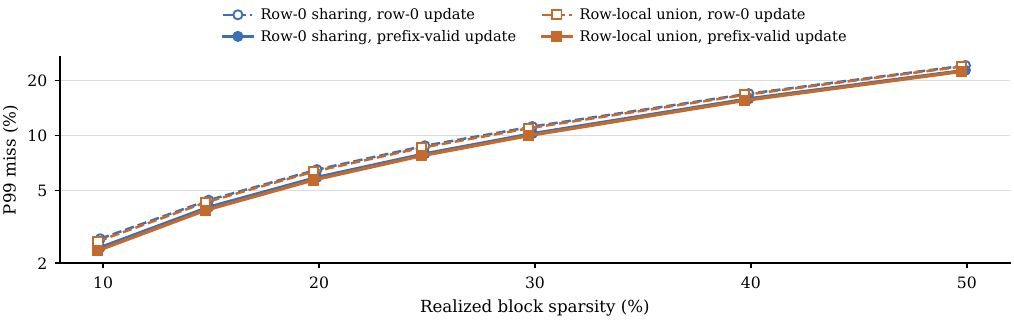}
    \caption{MTP policy ablation over 60 \bench{AA-LCR} traces with a draft length of three. Each point reports p99 miss and realized block sparsity for one target sparsity under one mask-construction and EMA-update strategy.}
    \label{fig:mtp-policy}
\end{figure*}

Figure~\ref{fig:mtp-policy} shows that prefix-valid update has the larger effect, reducing p99 miss by 5.9--11.0\% relative to row-0 update. Mask construction has a smaller effect. At matched realized sparsity, row-local union achieves slightly lower p99 miss than row-0 sharing. Mean miss follows the same trend and is therefore omitted from the figure. We therefore adopt row-local union with prefix-valid update as \spin's default MTP strategy.

\paragraph{Online acceptance behavior.}
We then evaluate the selected strategy online on \bench{SPEED-Bench} using \dsvfour Flash. We compare it with all-keep MTP and a sparse baseline using row-0 sharing with row-0 update. Both sparse configurations use 40\% target block sparsity. Average acceptance length is the mean number of tokens committed per verification step, including the target-model token, and therefore ranges from one to four. Draft-token acceptance rate is the fraction of proposed draft tokens accepted by the target model.

We exclude 27 responses flagged for repetition across all three algorithms (0.59--0.85\% per configuration) and ten requests with incomplete mask statistics, leaving 1,499 matched requests.

\begin{table}[h]
    \centering
    \caption{Online MTP acceptance behavior on a common subset of 1,499 \bench{SPEED-Bench} requests after excluding repetition-flagged responses and requests without complete mask statistics.}
    \label{tab:mtp-online}
    \begin{tabular*}{\linewidth}{l@{\extracolsep{\fill}}rrr}
        \toprule
        Condition &
        \shortstack{Average acceptance\\length} &
        \shortstack{Draft-token\\acceptance rate} &
        \shortstack{Realized\\sparsity} \\
        \midrule
        All-keep MTP & 2.4049 & 47.19\% & -- \\
        Row-0 sharing + row-0 update & 2.4059 & 47.39\% & 39.67\% \\
        Row-local union + prefix-valid update & 2.4097 & 47.28\% & 39.51\% \\
        \bottomrule
    \end{tabular*}
\end{table}

Across the three configurations, average acceptance length differs by at most 0.0049, and draft-token acceptance rate differs by less than 0.5\% relative to all-keep. The selected \spin strategy therefore shows no visible change in observed MTP acceptance behavior on this matched subset while retaining 39.51\% realized block sparsity. Its realized sparsity is only 0.4\% lower than that of the row-0-sharing baseline, indicating that verifier rows select largely overlapping KV blocks.

\subsection{\spin Design Ablations}

\paragraph{Block reduction.}
\label{sec:block-reduction}

We ablate two block-reduction design choices in \spin: where in the prediction algorithm slot-level signals are reduced to blocks and how values within each block are pooled. We define three variants, ordered from the latest to the earliest block reduction:
\begin{itemize}[nosep,leftmargin=*]
    \item \textbf{A0:} Compute predicted slot scores, select \topk slots, then expand the selection to blocks. Every block containing at least one selected slot is selected.
    \item \textbf{A1 (\spin's default):} Compute predicted slot scores, pool them within each block to obtain predicted block scores, and select the top-$m_t$ blocks.
    \item \textbf{A2:} Pool indexer scores within each block and use the pooled scores to update block-level $V/D$ EMA states. Use these states to directly predict block scores for top-$m_t$ block selection.
\end{itemize}
A0 therefore performs block conversion latest, after slot selection. A1 pools predicted slot scores into predicted block scores while retaining slot-level EMA states. A2 performs block reduction earliest, before the EMA predictor. A0 has no pooling operator. For A1 and A2, Max takes the largest value in each block, Top-4 averages up to the four largest values, and Mean averages all values.

\begin{figure*}[hbt]
    \centering
    \includegraphics[width=\textwidth]{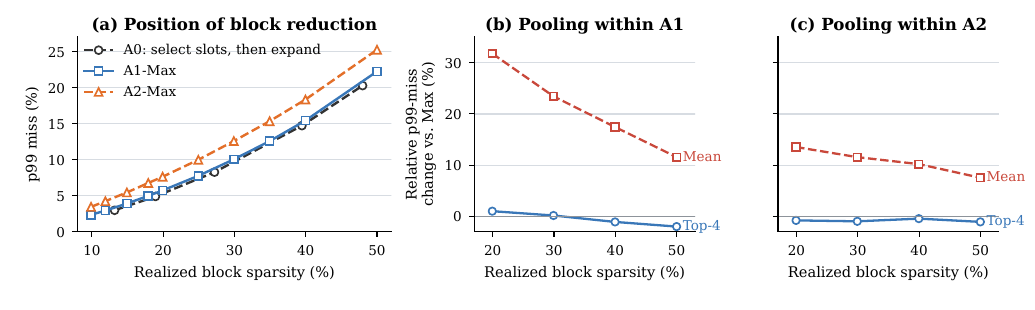}
    \caption{\textbf{Block-reduction ablation.} Panel (a) compares P99 miss rates for A0, A1-Max, and A2-Max; panels (b) and (c) show relative changes in P99 miss for Mean and Top-4 compared with Max pooling in A1 and A2. Results average per-layer, per-trace P99 over all 60 curated \bench{AA-LCR} traces and CSA layers.}
    \label{fig:pooling-ablation}
\end{figure*}

\begingroup
\widowpenalty=10000
Figure~\ref{fig:pooling-ablation}(a) shows that A0 and A1-Max remain close, whereas A2-Max produces consistently higher p99 miss. Mean pooling consistently increases p99 miss. Top-4 remains close to Max but is not uniformly better. Within A1, it is slightly worse at 10--20\% sparsity, near parity at 25--35\%, and modestly better at 40--50\%. We choose A1 because it directly controls target block sparsity, whereas A0 controls a pre-expansion slot budget and therefore determines realized block sparsity only indirectly. A1-Max also achieves lower p99 miss than A2-Max. We choose Max over Top-4 because it is simpler and avoids the low-sparsity regression while remaining close across the evaluated range.
\par\endgroup

\paragraph{Exploration methods.}
\phantomsection\label{sec:staleness}

The MRCRv2 results show that random exploration recovers task quality without increasing the selected-block budget. We compare it with two alternatives under the same setup. Staleness-score exploration combines staleness and predicted importance to rank candidate blocks. Block-age exploration selects the least recently observed blocks. All three methods use a 5\% exploration share.

\begin{figure*}[!ht]
    \centering
    \includegraphics[width=\textwidth]{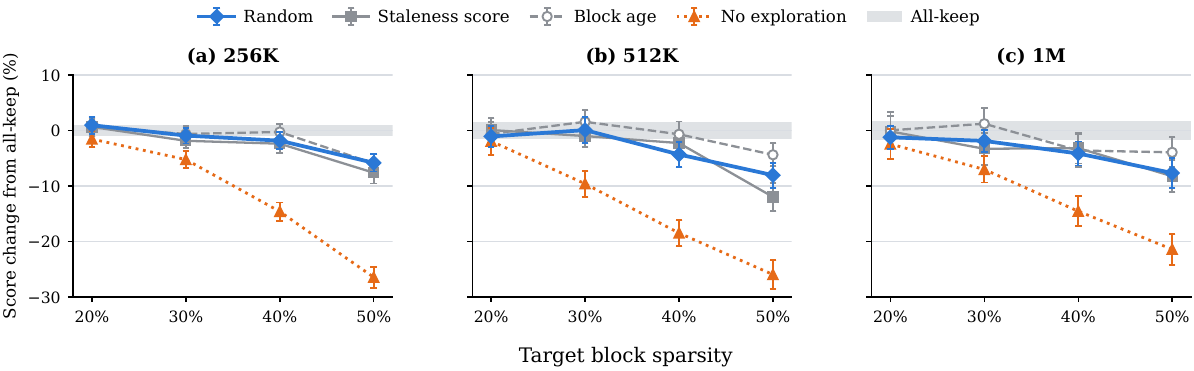}
    \caption{Exploration-method comparison on \bench{MRCRv2} at 256K, 512K, and 1M context lengths for \dsvfour Flash. Each point is the mean score change from all-keep, in percent, over 40 repeats of 100 fixed prompts. Error bars are 95\% confidence intervals of the paired difference, and the gray band is the all-keep 95\% confidence interval. All exploration methods use a 5\% exploration share.}
    \label{fig:exploration-policies}
\end{figure*}

All three methods recover most of the quality lost without exploration, so the choice of which skipped blocks to refresh is secondary. \spin uses random exploration, even though block-age exploration scores higher at 512K and 1M (Figure~\ref{fig:exploration-policies}). Random exploration needs no age tracking or age-based selection.

\begin{figure*}[!ht]
    \centering
    \includegraphics[width=\textwidth]{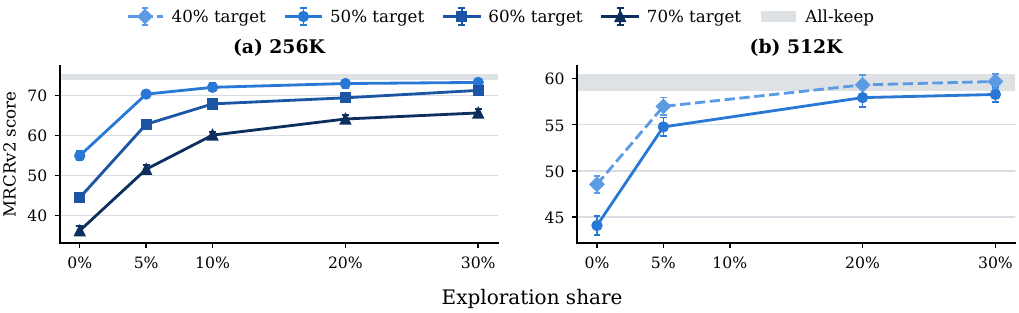}
    \caption{MRCRv2 task score versus exploration share for \dsvfour Flash with random exploration. (a) 256K at 50\%, 60\%, and 70\% target block sparsity. (b) 512K at 40\% and 50\% target block sparsity with 0--30\% exploration shares. Each point averages 40 repeats over 100 fixed prompts. Error bars are 95\% confidence intervals of the mean, and gray bands are all-keep 95\% confidence intervals.}
    \label{fig:exploration-share}
\end{figure*}

\paragraph{Exploration share.}
Higher target block sparsity needs a larger exploration share (Figure~\ref{fig:exploration-share}). With a 20--30\% share, task quality stays close to all-keep at 50\% target block sparsity. Because the share does not change throughput (Section~\ref{sec:end-to-end-performance}), scaling it with block sparsity is free.

\begin{figure}[!ht]
    \centering
    \includegraphics[width=0.85\textwidth]{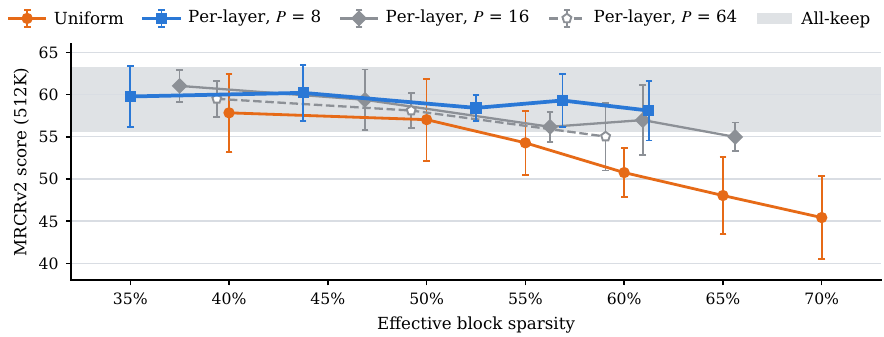}
    \caption{MRCRv2 task score versus effective block sparsity at 512K for \dsvfour Flash, with a 5\% exploration share. $P$ is the probe period, and effective block sparsity counts probe steps as dense. Each point averages 5 repeats over 100 fixed prompts. Error bars are 95\% confidence intervals of the mean, and the gray band is the 95\% confidence interval of an all-keep control run with the same repeats.}
    \label{fig:per-layer-quality}
\end{figure}

\paragraph{Per-layer block budgets.}
\label{sec:per-layer-budget-results}
We evaluate per-layer block budgets on 512K \bench{MRCRv2} with the deployment of Section~\ref{sec:mrcr} and the default 5\% exploration share. With a probe period of 8, they keep task quality close to all-keep at 57\% and 61\% effective block sparsity, 5--7 points above a uniform budget at 55\% and 60\% (Figure~\ref{fig:per-layer-quality}).

\spin sizes a per-layer block budget, keeping the total budget across all layers at the same user-specified target block sparsity. The allocation minimizes the worst layer's miss rate, not the average. Empirically, this objective performs better, scoring about 1.28 points higher on average than minimizing the average layer's miss rate across 50--70\% target sparsity (the Pareto points in Figure~\ref{fig:per-layer-quality}).

\FloatBarrier
\section{Related Work}
\label{sec:related}

\paragraph{Temporal correlation in attention.}
It is well documented that attention scores form vertical and diagonal patterns in certain heads; many papers have leveraged these patterns to sparsify and speed up attention during decode. H$_2$O~\cite{h2o2023} selects heavy-hitter tokens using accumulated historical attention scores while retaining a separate budget for recent tokens. StreamingLLM~\cite{streamingllm2024} instead retains attention-sink tokens together with a sliding window of recent tokens. Several works also analyze the diagonal pattern and its origin in RoPE positional encoding~\cite{barbero2025round,slashrope2026,tappa2026}. AttentionPredictor~\cite{attnpredictor2025} learns to predict next-token attention scores from their temporal history. PRR~\cite{prr2026} applies an EMA block predictor to dynamic sparse attention, using predicted blocks to speculatively execute core attention while the original block selector runs and repairing missed blocks afterward. Rather than assuming temporal prediction is the right choice, Section~\ref{sec:patterns} empirically compares previous-iteration prediction against random and cross-layer alternatives and shows that previous-iteration prediction is more accurate.

\paragraph{Indexer-based sparse attention and its optimizations.}
Recent work reduces indexer cost through several complementary mechanisms. Cross-layer reuse avoids repeated indexing at adjacent layers. IndexCache~\cite{indexcache2026}, discussed in Section~\ref{sec:patterns}, studies training-free and training-aware methods that reuse \topk indices across DSA layers. GLM-5.2 natively deploys this mechanism from mid-training~\cite{glm52}. HISA~\cite{hisa2026} reduces indexer cost through hierarchical filtering, running DSA's original token-level indexer only within blocks selected by a coarse block-level stage. LongCat-2.0 later combines cross-layer reuse and hierarchical filtering with streaming-aware indexing in LongCat Sparse Attention~\cite{longcat2026}. MiniMax Sparse Attention~\cite{minimaxmsa2026} instead adopts a native block-level Index Branch that selects blocks independently for each GQA group. MISA~\cite{misa2026} routes a query-dependent subset of indexer heads to perform full-prefix token-level scoring, while PIVOT~\cite{pivot2026} amortizes a full-prefix indexer scan across grouped queries. Closest to our work, TISA~\cite{tisa2026} partitions prefix positions into fixed contiguous groups, refreshes one group per decode step in round-robin order, and reuses cached scores for the remaining groups before running \topk over the complete score cache. Unlike TISA's fixed refresh schedule, \spin explicitly models vertical and diagonal temporal patterns with per-slot and relative-offset EMA states and selects indexer-input blocks according to predicted importance.

\paragraph{Non-uniform budget allocation.}
KV-cache compression methods have also found that a uniform budget is suboptimal. PyramidKV~\cite{pyramidkv2024} assigns larger cache budgets to lower layers, following a pyramidal pattern of attention concentration. Ada-KV~\cite{adakv2024} redistributes a fixed budget across the heads of a layer according to their attention scores. Both methods permanently evict KV entries and set their allocation from attention statistics. \spin's per-layer block budget keeps the full cache and controls which blocks enter the indexer at each decode step. It allocates blocks to minimize the worst-layer miss of block prediction against the dense indexer, measured by periodic dense probes and refit online as the workload drifts (Section~\ref{sec:per-layer-budget}).

Other work targets later stages of sparse attention. Guess-Verify-Refine~\cite{gvr2026} uses temporal correlation to accelerate the post-indexer \topk operator. \spin instead applies temporal prediction before the indexer, using vertical and diagonal slot-level EMAs to prune its input, and refreshes skipped blocks with exploration within the selected-block budget.

\section{Limitations}
\label{sec:limitations}

\spin predicts block importance from historical indexer scores and therefore has limited ability to anticipate patterns not represented in its history, such as sudden attention shifts. Incorporating other prediction signals is left for future work.

Our current evaluation focuses on \dsvfour and its native MTP. Although we expect the observations and designs to apply more broadly, this has not been demonstrated on other indexer-based sparse-attention models, such as GLM-5, MiniMax-M3, and LongCat-2.0, or with other speculative-decoding methods.

\section{Conclusion}
\label{sec:conclusion}

\spin shows that temporal statistics of indexer scores can serve as an effective proxy for current block importance. Using vertical and diagonal EMA statistics, \spin skips up to 40\% of indexer input blocks while preserving comparable task quality across the evaluated long-context and agentic benchmarks. Random exploration keeps observations fresh within the same block budget. On \bench{MRCRv2}, exploration shares that grow with target block sparsity and per-layer block budgets keep task quality close to all-keep at higher block sparsity. \spin also retains block sparsity without a visible change in acceptance behavior under \dsvfour's native MTP. At 57\% effective block sparsity, our integrated vLLM implementation improves output throughput by 13.9\% and reduces median inter-token latency by 13.6\%.  {}
Combining temporal history with current-query or cross-layer signals may reduce worst-case misses under sudden attention shifts and is a natural direction for future work.

\bibliographystyle{plainnat}
\bibliography{references}

\begin{thebibliography}{32}
\providecommand{\natexlab}[1]{#1}
\providecommand{\url}[1]{\texttt{#1}}
\expandafter\ifx\csname urlstyle\endcsname\relax
  \providecommand{\doi}[1]{doi: #1}\else
  \providecommand{\doi}{doi: \begingroup \urlstyle{rm}\Url}\fi

\bibitem[Abramovich et~al.(2026)Abramovich, Ashkenazi, Putterman, Chislett,
  Mitra, Rouhani, Zilberstein, and Geifman]{speedbench2026}
Talor Abramovich, Maor Ashkenazi, Izzy Putterman, Benjamin Chislett, Tiyasa
  Mitra, Bita~Darvish Rouhani, Ran Zilberstein, and Yonatan Geifman.
\newblock {SPEED-Bench}: A unified and diverse benchmark for speculative
  decoding.
\newblock \emph{arXiv preprint}, February 2026.
\newblock URL \url{https://arxiv.org/abs/2604.09557}.

\bibitem[{Anonymous}(2026)]{tisa2026}
{Anonymous}.
\newblock {TISA}: Exploiting temporal locality to accelerate indexing sparse
  attention.
\newblock \emph{ACL Rolling Review (ARR) submission}, May 2026.
\newblock URL \url{https://openreview.net/forum?id=6kOo3YtMdu}.

\bibitem[{Artificial Analysis}(2025)]{artificialanalysis2025lcr}
{Artificial Analysis}.
\newblock Artificial analysis long context reasoning benchmark leaderboard,
  2025.
\newblock URL
  \url{https://artificialanalysis.ai/evaluations/artificial-analysis-long-context-reasoning}.

\bibitem[Bai et~al.(2025)Bai, Tu, Zhang, Peng, Wang, Lv, Cao, Xu, Hou, Dong,
  Tang, and Li]{bai2025longbenchv2deeperunderstanding}
Yushi Bai, Shangqing Tu, Jiajie Zhang, Hao Peng, Xiaozhi Wang, Xin Lv, Shulin
  Cao, Jiazheng Xu, Lei Hou, Yuxiao Dong, Jie Tang, and Juanzi Li.
\newblock Longbench v2: Towards deeper understanding and reasoning on realistic
  long-context multitasks, 2025.
\newblock URL \url{https://arxiv.org/abs/2412.15204}.

\bibitem[Bai et~al.(2026)Bai, Dong, Jiang, Lv, Du, Zeng, Tang, and
  Li]{indexcache2026}
Yushi Bai, Qian Dong, Ting Jiang, Xin Lv, Zhengxiao Du, Aohan Zeng, Jie Tang,
  and Juanzi Li.
\newblock {IndexCache}: Accelerating sparse attention via cross-layer index
  reuse.
\newblock \emph{arXiv preprint}, March 2026.
\newblock URL \url{https://arxiv.org/abs/2603.12201}.

\bibitem[Barbero et~al.(2025)Barbero, Vitvitskyi, Perivolaropoulos, Pascanu,
  and Veli{\v{c}}kovi{\'c}]{barbero2025round}
Federico Barbero, Alex Vitvitskyi, Christos Perivolaropoulos, Razvan Pascanu,
  and Petar Veli{\v{c}}kovi{\'c}.
\newblock Round and round we go! what makes rotary positional encodings useful?
\newblock In \emph{The Thirteenth International Conference on Learning
  Representations}, 2025.
\newblock URL \url{https://openreview.net/forum?id=GtvuNrk58a}.

\bibitem[Barres et~al.(2025)Barres, Dong, Ray, Si, and
  Narasimhan]{barres2025tau2}
Victor Barres, Honghua Dong, Soham Ray, Xujie Si, and Karthik Narasimhan.
\newblock $\tau^2$-bench: Evaluating conversational agents in a dual-control
  environment, 2025.
\newblock URL \url{https://arxiv.org/abs/2506.07982}.

\bibitem[Cai et~al.(2024)Cai, Zhang, Gao, Liu, Liu, Lu, Xiong, Dong, Chang, Hu,
  and Xiao]{pyramidkv2024}
Zefan Cai, Yichi Zhang, Bofei Gao, Yuliang Liu, Tianyu Liu, Keming Lu, Wayne
  Xiong, Yue Dong, Baobao Chang, Junjie Hu, and Wen Xiao.
\newblock {PyramidKV}: Dynamic {KV} cache compression based on pyramidal
  information funneling.
\newblock \emph{arXiv preprint arXiv:2406.02069}, 2024.

\bibitem[Chen et~al.(2022)Chen, Chu, Wiseman, and Gimpel]{summscreen2022}
Mingda Chen, Zewei Chu, Sam Wiseman, and Kevin Gimpel.
\newblock {SummScreen}: A dataset for abstractive screenplay summarization.
\newblock \emph{Annual Meeting of the Association for Computational Linguistics
  (ACL)}, May 2022.
\newblock URL \url{https://aclanthology.org/2022.acl-long.589}.

\bibitem[Cheng et~al.(2026{\natexlab{a}})Cheng, Zhao, Liu, Li, Qiao, Duan,
  Chen, Chen, Rouhani, and Yang]{gvr2026}
Long Cheng, Ritchie Zhao, Timmy Liu, Mindy Li, Xianjie Qiao, Kefeng Duan,
  Yu-Jung Chen, Xiaoming Chen, Bita~Darvish Rouhani, and June Yang.
\newblock {Guess-Verify-Refine}: Data-aware top-{K} for sparse-attention
  decoding on {Blackwell} via temporal correlation.
\newblock \emph{arXiv preprint}, April 2026{\natexlab{a}}.
\newblock URL \url{https://arxiv.org/abs/2604.22312}.

\bibitem[Cheng et~al.(2026{\natexlab{b}})Cheng, Zhang, Hou, Du, Du, Pang, Sun,
  and Yang]{slashrope2026}
Yuan Cheng, Fengzhuo Zhang, Yunlong Hou, Cunxiao Du, Chao Du, Tianyu Pang,
  Aixin Sun, and Zhuoran Yang.
\newblock Demystifying the slash pattern in attention: The role of {RoPE}.
\newblock \emph{arXiv preprint}, 2026{\natexlab{b}}.
\newblock URL \url{https://arxiv.org/abs/2601.08297}.

\bibitem[DeepSeek-AI(2025)]{deepseekv32}
DeepSeek-AI.
\newblock {DeepSeek-V3.2}: Pushing the frontier of open large language models.
\newblock \emph{arXiv preprint}, December 2025.
\newblock URL \url{https://arxiv.org/abs/2512.02556}.

\bibitem[DeepSeek-AI(2026)]{deepseekv4}
DeepSeek-AI.
\newblock {DeepSeek-V4}: Towards highly efficient million-token context
  intelligence.
\newblock \emph{arXiv preprint}, April 2026.
\newblock URL \url{https://arxiv.org/abs/2606.19348}.

\bibitem[Feng et~al.(2024)Feng, Lv, Cao, Xie, and Zhou]{adakv2024}
Yuan Feng, Junlin Lv, Yukun Cao, Xike Xie, and S.~Kevin Zhou.
\newblock {Ada-KV}: Optimizing {KV} cache eviction by adaptive budget
  allocation for efficient {LLM} inference.
\newblock \emph{arXiv preprint arXiv:2407.11550}, 2024.

\bibitem[Hsieh et~al.(2024)Hsieh, Sun, Kriman, Acharya, Rekesh, Jia, Zhang, and
  Ginsburg]{ruler2024}
Cheng-Ping Hsieh, Simeng Sun, Samuel Kriman, Shantanu Acharya, Dima Rekesh, Fei
  Jia, Yang Zhang, and Boris Ginsburg.
\newblock {RULER}: What's the real context size of your long-context language
  models?
\newblock \emph{Conference on Language Modeling (COLM)}, April 2024.
\newblock URL \url{https://arxiv.org/abs/2404.06654}.

\bibitem[Jimenez et~al.(2024)Jimenez, Yang, Wettig, Yao, Pei, Press, and
  Narasimhan]{swebench2024}
Carlos~E. Jimenez, John Yang, Alexander Wettig, Shunyu Yao, Kexin Pei, Ofir
  Press, and Karthik Narasimhan.
\newblock {SWE-bench}: Can language models resolve real-world {GitHub} issues?
\newblock In \emph{The Twelfth International Conference on Learning
  Representations}, 2024.
\newblock URL \url{https://openreview.net/forum?id=VTF8yNQM66}.

\bibitem[Kwon et~al.(2023)Kwon, Li, Zhuang, Sheng, Zheng, Yu, Gonzalez, Zhang,
  and Stoica]{vllm2023}
Woosuk Kwon, Zhuohan Li, Siyuan Zhuang, Ying Sheng, Lianmin Zheng, Cody~Hao Yu,
  Joseph~E. Gonzalez, Hao Zhang, and Ion Stoica.
\newblock Efficient memory management for large language model serving with
  {PagedAttention}.
\newblock \emph{Symposium on Operating Systems Principles (SOSP)}, September
  2023.
\newblock URL \url{https://arxiv.org/abs/2309.06180}.

\bibitem[Lai et~al.(2026)Lai, Xu, Yang, Chen, Xu, Zeng, Li, Sun, Zhu, Zhang,
  Hu, Li, Gao, Li, Zhu, Zhou, and Zhao]{minimaxmsa2026}
Xunhao Lai, Weiqi Xu, Yufeng Yang, Qiaorui Chen, Yang Xu, Lunbin Zeng, Xiaolong
  Li, Haohai Sun, Haichao Zhu, Vito Zhang, Jinkai Hu, Jiayao Li, Rui Gao, Zekun
  Li, Songquan Zhu, Jingkai Zhou, and Pengyu Zhao.
\newblock {MiniMax} sparse attention.
\newblock \emph{arXiv preprint}, June 2026.
\newblock URL \url{https://arxiv.org/abs/2606.13392}.

\bibitem[Liu et~al.(2026)Liu, Cheng, Niu, Su, Xue, Liu, Yu, and Zhu]{pivot2026}
Hong Liu, Yuan Cheng, Lin Niu, Yi~Su, Yufei Xue, Anmin Liu, Guanghua Yu, and
  Jianchen Zhu.
\newblock {PIVOT}: Efficient query-group indexing for token-level sparse
  attention.
\newblock \emph{arXiv preprint}, July 2026.
\newblock URL \url{https://arxiv.org/abs/2607.24593}.

\bibitem[{LongCat}(2026)]{longcat2026}
{LongCat}.
\newblock Introducing {LongCat-2.0}, June 2026.
\newblock URL \url{https://longcat.chat/blog/longcat-2.0/}.

\bibitem[{OpenAI}(2025)]{openai2025mrcr}
{OpenAI}.
\newblock Introducing {GPT-4.1} in the {API}, April 2025.
\newblock URL \url{https://openai.com/index/gpt-4-1/}.

\bibitem[Su et~al.(2021)Su, Lu, Pan, Murtadha, Wen, and Liu]{rope2021}
Jianlin Su, Yu~Lu, Shengfeng Pan, Ahmed Murtadha, Bo~Wen, and Yunfeng Liu.
\newblock {RoFormer}: Enhanced transformer with rotary position embedding.
\newblock \emph{arXiv preprint}, April 2021.
\newblock URL \url{https://arxiv.org/abs/2104.09864}.

\bibitem[Wang et~al.(2026)Wang, Rattihalli, Dhakal, Li, Ren, Milojicic, and
  Shangguan]{prr2026}
Tianyu Wang, Gourav Rattihalli, Aditya Dhakal, Junbo Li, Zhiwei Ren, Dejan
  Milojicic, and Longfei Shangguan.
\newblock Predict, reuse, and repair: Accelerating dynamic sparse attention for
  long-context {LLM} decoding.
\newblock \emph{arXiv preprint}, June 2026.
\newblock URL \url{https://arxiv.org/abs/2606.30389}.

\bibitem[Xiao et~al.(2024)Xiao, Tian, Chen, Han, and Lewis]{streamingllm2024}
Guangxuan Xiao, Yuandong Tian, Beidi Chen, Song Han, and Mike Lewis.
\newblock Efficient streaming language models with attention sinks.
\newblock \emph{International Conference on Learning Representations (ICLR)},
  September 2024.
\newblock URL \url{https://arxiv.org/abs/2309.17453}.

\bibitem[Xu et~al.(2026)Xu, Meng, Jiang, Wang, Zhou, Wang, Wu, Pan, Tang, Pei,
  Liu, Yin, Sun, and Zhang]{hisa2026}
Yufei Xu, Fanxu Meng, Fan Jiang, Yuxuan Wang, Ruijie Zhou, Zhaohui Wang, Jiexi
  Wu, Zhixin Pan, Xiaojuan Tang, Wenjie Pei, Tongxuan Liu, Di~Yin, Xing Sun,
  and Muhan Zhang.
\newblock {HISA}: Efficient hierarchical indexing for fine-grained sparse
  attention.
\newblock \emph{arXiv preprint}, March 2026.
\newblock URL \url{https://arxiv.org/abs/2603.28458}.

\bibitem[Yang et~al.(2025)Yang, Wang, Li, Wang, Chen, Chen, Yu, Liu, Hao, Yuan,
  and Li]{attnpredictor2025}
Qingyue Yang, Jie Wang, Xing Li, Zhihai Wang, Chen Chen, Lei Chen, Xianzhi Yu,
  Wulong Liu, Jianye Hao, Mingxuan Yuan, and Bin Li.
\newblock {AttentionPredictor}: Temporal patterns matter for {KV} cache
  compression.
\newblock \emph{Conference on Neural Information Processing Systems (NeurIPS)},
  February 2025.
\newblock URL \url{https://arxiv.org/abs/2502.04077}.

\bibitem[Yang et~al.(2026)Yang, Wang, Li, Bai, Tong, Zhen, Hao, Yuan, and
  Li]{tappa2026}
Qingyue Yang, Jie Wang, Xing Li, Yinqi Bai, Xialiang Tong, Huiling Zhen, Jianye
  Hao, Mingxuan Yuan, and Bin Li.
\newblock Why attention patterns exist: A unifying temporal perspective
  analysis.
\newblock \emph{International Conference on Learning Representations (ICLR)},
  2026.
\newblock URL \url{https://arxiv.org/abs/2601.21709}.

\bibitem[Yao et~al.(2024)Yao, Shinn, Razavi, and Narasimhan]{yao2024tau}
Shunyu Yao, Noah Shinn, Pedram Razavi, and Karthik Narasimhan.
\newblock $\tau$-bench: A benchmark for tool-agent-user interaction in
  real-world domains, 2024.
\newblock URL \url{https://arxiv.org/abs/2406.12045}.

\bibitem[{Z.ai}(2026)]{glm52}
{Z.ai}.
\newblock {GLM-5.2}: Built for long-horizon tasks, June 2026.
\newblock URL \url{https://z.ai/blog/glm-5.2}.

\bibitem[Zhang et~al.(2023)Zhang, Sheng, Zhou, Chen, Zheng, Cai, Song, Tian,
  R{\'e}, Barrett, Wang, and Chen]{h2o2023}
Zhenyu Zhang, Ying Sheng, Tianyi Zhou, Tianlong Chen, Lianmin Zheng, Ruisi Cai,
  Zhao Song, Yuandong Tian, Christopher R{\'e}, Clark Barrett, Zhangyang Wang,
  and Beidi Chen.
\newblock {H2O}: Heavy-hitter oracle for efficient generative inference of
  large language models.
\newblock \emph{Conference on Neural Information Processing Systems (NeurIPS)},
  June 2023.
\newblock URL \url{https://arxiv.org/abs/2306.14048}.

\bibitem[Zhao et~al.(2024)Zhao, Ren, Hessel, Cardie, Choi, and
  Deng]{wildchat2024}
Wenting Zhao, Xiang Ren, Jack Hessel, Claire Cardie, Yejin Choi, and Yuntian
  Deng.
\newblock {WildChat}: 1{M} {ChatGPT} interaction logs in the wild.
\newblock In \emph{The Twelfth International Conference on Learning
  Representations}, 2024.
\newblock URL \url{https://arxiv.org/abs/2405.01470}.

\bibitem[Zhou et~al.(2026)Zhou, Meng, Xu, Liu, Lu, Zhang, and Pei]{misa2026}
Ruijie Zhou, Fanxu Meng, Yufei Xu, Tongxuan Liu, Guangming Lu, Muhan Zhang, and
  Wenjie Pei.
\newblock {MISA}: Mixture of indexer sparse attention for long-context {LLM}
  inference.
\newblock \emph{arXiv preprint}, May 2026.
\newblock URL \url{https://arxiv.org/abs/2605.07363}.

\end{thebibliography}

\end{document}